\pdfoutput=1

\documentclass[11pt]{article}

\usepackage[]{EMNLP2022}

\usepackage{times}
\usepackage{latexsym}
\usepackage{graphicx}
\usepackage{amsmath}
\usepackage{amssymb}
\usepackage{booktabs}
\usepackage{textcomp}
\usepackage{multirow}
\usepackage{array}
\usepackage{bbm}
\usepackage{tabularx, booktabs}
\newcolumntype{Y}{>{\centering\arraybackslash}X}
\newcolumntype{L}{>{\arraybackslash}X}

\usepackage{inconsolata}
\usepackage{tablefootnote}

\usepackage[T1]{fontenc}

\usepackage[utf8]{inputenc}

\usepackage{microtype}

\usepackage{xparse}
\definecolor{bluepigment}{rgb}{0.2, 0.2, 0.6}

\newcommand{\ours}{\textsc{TAbs}}

\title{Open Relation and Event Type Discovery with Type Abstraction} 

\author{Sha Li, Heng Ji, Jiawei Han \\
University of Illinois Urbana-Champaign \\
\texttt{\{shal2, hengji, hanj\}@illinois.edu} 
}

\begin{document}
\maketitle
\begin{abstract}
Conventional ``closed-world" information extraction (IE) approaches rely on human ontologies to define the scope for extraction. As a result, such approaches fall short when applied to new domains. This calls for systems that can automatically infer new types from given corpora, a task which we refer to as \textit{type discovery}.
To tackle this problem, we introduce the idea of type abstraction, where the model is prompted to generalize and name the type. Then we use the similarity between inferred names to induce clusters. Observing that this abstraction-based representation is often complementary to the entity/trigger token representation, we set up these two representations as two views and design our model as a co-training framework. 
Our experiments on multiple relation extraction and event extraction datasets consistently show the advantage of our type abstraction approach.

\end{abstract}
\section{Introduction}
Information extraction has enjoyed widespread success, however, 
the majority of information extraction methods are ``reactive'', relying on end-users to specify their information needs in prior and provide supervision accordingly.  This leads to ``closed-world'' systems~\cite{lin-etal-2020-joint,du-cardie-2020-event,li-etal-2021-document,zhong-chen-2021-frustratingly,ye-etal-2022-packed} that are confined to a set of pre-defined types. %
It is desirable to make systems act more ``proactively'' like humans who are always on the lookout for interesting new information, generalize them into new types, and find more instances of such types, even if they are not seen previously. 

One related attempt is the Open Information Extraction paradigm~\cite{Banko2008OpenIE}, which aims at extracting all (subject, predicate, object) triples from text that denote some kind of relation. While OpenIE does not rely on pre-specified relations, its exhaustive and free-form nature 
often leads to noisy and redundant extractions.

To bridge the gap between closed-world IE and OpenIE, a vital step is for systems to possess the ability of automatically inducing new types and extracting instances of such new types. 
Under various contexts, related methods have been proposed under the name of ``relation discovery''~\cite{yao-etal-2011-structured, marcheggiani-titov-2016-discrete},``open relation extraction''~\cite{wu-etal-2019-open,hu-etal-2020-selfore} and ``event type induction''~\cite{huang-ji-2020-semi, shen-etal-2021-corpus}. In this paper, we unify such terms and refer to the task as \textit{type discovery}.
\begin{figure}[t]
    \centering
    \includegraphics[width=.9\linewidth]{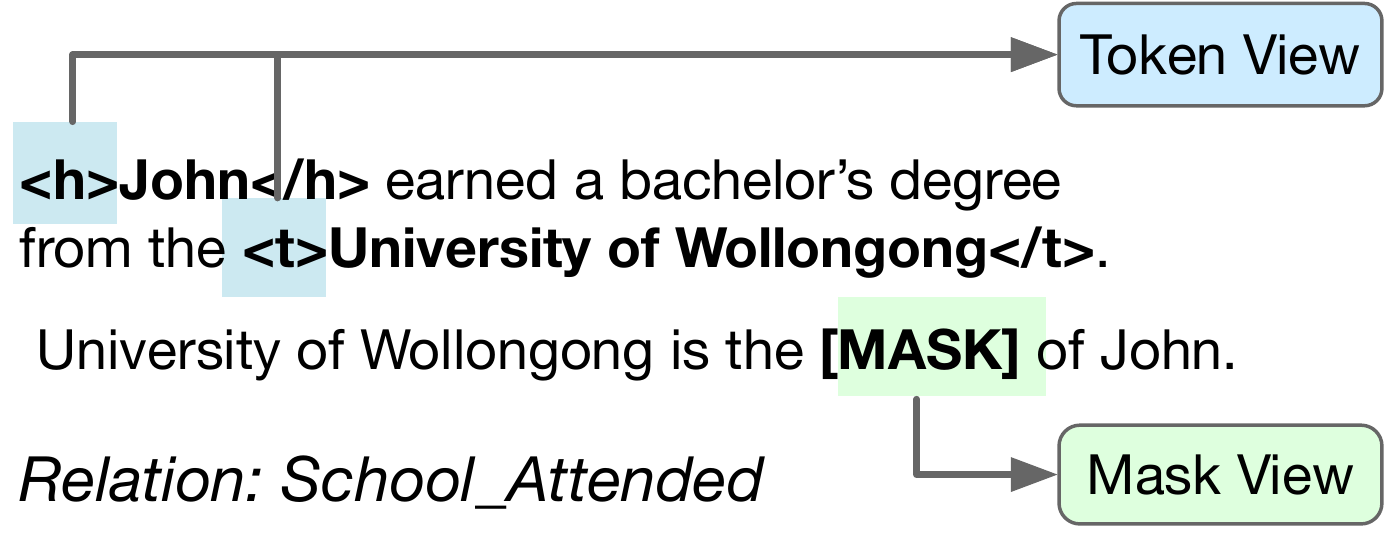}
    \caption{For each instance, the token view is computed from the pre-trained LM embedding of the first token in entity/trigger. The mask view is computed from the \texttt{[MASK]} token embedding in the type prompt.}
    \label{fig:two_view_example}
\end{figure}

\textit{Type discovery} can naturally be posed as a clustering task.
This heavily relies on defining an appropriate metric space where types are easily separable.
The token embedding space from pre-trained language models is a popular choice, but 
as observed by ~\cite{zhao-etal-2021-relation}, the original metric space derived from BERT~\cite{devlin-etal-2019-bert} is often prone to reflect surface form similarity rather than the desired relation/event-centered similarity.
One way to alleviate this issue is to 
use known types to help learn a similarity metric that can also be applied to unknown types ~\cite{wu-etal-2019-open,zhao-etal-2021-relation,huang-ji-2020-semi}.

\begin{table}[ht]
    \centering
    \small 
    \begin{tabular}{l| c c  c}
    \toprule 
        Relation & Mask view & Token view & $\Delta$ \\
    \midrule 
      website (of org) & 0.2424 &  0.9366 & -0.6941 \\
     age (of person) & 0.2896  & 0.389  & -0.0994 \\
      founded\_by & 0.2734 & 0.1268  & 0.1466 \\
    employee\_of & 0.4434 & 0.2703  & 0.1731 \\
     \midrule 
     Avg &0.3678 & 0.2989 & 0.0688 \\
       \bottomrule 
    \end{tabular}
    \caption{Probing $k$-NN Accuracy of the token view and the mask view on distinguishing relations without training. We compute $k$-NN using cosine similarity of embeddings with $k=32$ on TACRED~\cite{zhang-etal-2017-position}. While on average the mask view outperforms the token view, the two views excel at different types.
    }
    \label{tab:knn_probe}
\end{table}

In this paper we introduce another idea of \textit{abstraction}: a discovered type should have an appropriate and concise type name. The human vocabulary serves as a good repository of concepts that appear meaningful to people. When we assign a name to a cluster, we implicitly define the commonality of instances within the cluster and also the criteria for including new instances to the cluster.
Since masked language models have the ability to ``fill in the blank'', with the help of a \textit{type-inducing prompt} as shown in Figure \ref{fig:two_view_example}, we can guide the model to predict a name or indicative word for any relation/event instance.
Moreover, since inferring the best name for a cluster from a single instance is a difficult task, we do not require this prediction to be exact: we utilize the similarity between predicted names to perform clustering.

This abstraction-based representation is complementary to the widely-adopted token-based representation of relations/events. 
We refer to our abstraction-based representation as ``mask view'' since the embedding for the instance is derived from the [MASK] token. Alternatively, we can also compute a ``token view'' derived from the pre-trained LM embeddings of %
the involved entity/trigger directly.
As shown in Table \ref{tab:knn_probe}, without any training, the token-based representation~(token view) and the type abstraction representation~(mask view) specialize in different types. When the relation type is strongly connected to the entity type as in ``website", the token view provides a strong prior. The mask view can distinguish relations with similar entity types (person, organization) based on relational phrases such as ``found, create, work at".

\begin{figure*}[t]
    \centering
    \includegraphics[width=\linewidth]{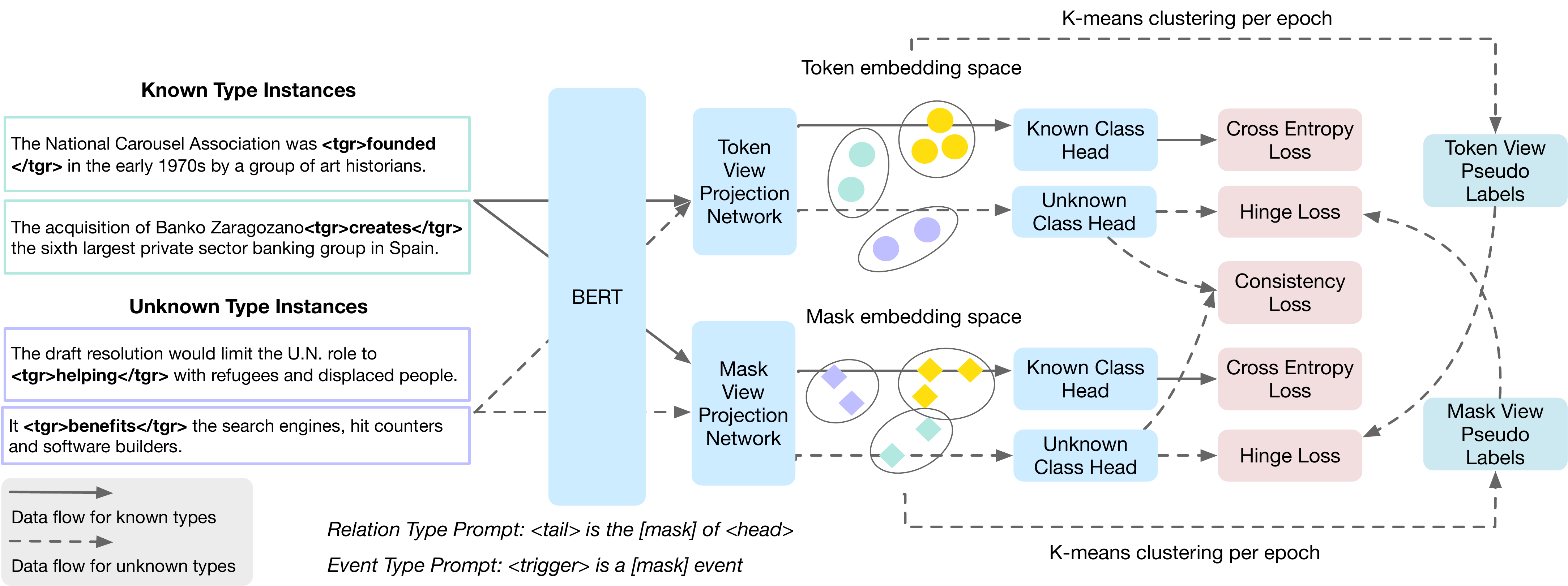}
    \caption{Overview of our type discovery model. We show two instances of the \texttt{Start-Org} event type (green) and two instances of \texttt{Assist} event type (purple). For each instance, we compute the token view and the mask view through two separate projection networks.  We use K-means clustering in the respective embedding spaces to obtain pseudo labels and use the labels to supervise the alternative view. 
    }
    \label{fig:framework}
\end{figure*}

Therefore, we combine the mask view and the token view in a co-training framework~\cite{Blum1998CombiningLA}, utilizing information from both ends. 
As shown in Figure \ref{fig:framework}, our model consists of a shared contextual encoder, two view-specific projection networks 
and classification layers for known and unknown types respectively.
Since no annotation is available for new types, we perform clustering over the two views to obtain pseudo-labels and then use such labels to guide the training of the classification layer of the opposite view. %

We apply our model to both relation discovery and event discovery with minimal changes to the type-inducing prompt. Our model serves the dual proposes of (1) inducing clusters with exemplar instances from the input corpus to assist ontology construction and (2) serving as a classifier for instances of unknown types.
On the task of relation discovery our model outperforms the previous transfer-learning based SOTA model by 4.3\% and 2.2\% accuracy on benchmark datasets TACRED and FewRel respectively. On event discovery we also set the new SOTA, achieving 77.3\% accuracy for type discovery with gold-standard triggers.

The main contributions of this paper include: 

\begin{itemize}
    \item We propose the idea of type abstraction, implicitly using inferred type names from the language model to improve type discovery.
    \item We design a co-training framework that combines the advantage of type abstraction and the conventional token-based representation.
    \item We show that our model can be applied to the discovery of both relation types and event types and achieve superior performance over existing models on both tasks. 
\end{itemize}

\section{Problem Definition}
We first define the task of \textbf{type discovery} and then discuss the realization of this task to relations and events.

Given a set of unlabeled instances $D^u = \{x_1^u, x_2^u, \cdots, x_M^u\}$ and an estimated number of unknown types $|C^u|$, the goal of type discovery is to learn a model $f$ that can map $x \in X^u$ into one of $y \in C^u$ unknown types.  

In the case of relation discovery, each instance $x$ is an entity mention pair $\{h, t\}$ embedded within a sentence context. As shown in Figure \ref{fig:two_view_example}, the instance is ``\textbf{John} earned a bachelor's degree from the \textbf{University of Wollongong}'' with the head entity mention ``John'' and the tail entity mention ``University of Wollongong''. Each entity mention is a span with start and end indexes $(s, e)$ in the sentence. The associated label $y$ in this case is relation type ``School\_Attended''.\footnote{In the relation extraction literature, the relation type is often denoted as $r$. For unified notation we use $y$.}

In the case of event discovery, each instance $x$ is a trigger word/phrase  mention $t$ with start and end indexes $(s, e)$ in a sentence context as shown in Figure \ref{fig:framework}. The label $y$ is the event type. 
Note that for both relations and events, it is possible for multiple instances to appear within the same sentence, but they have different entity or trigger mentions.

To assist the learning of such a model, we further assume that we have access to a set of labeled instances $D^l = \{(x_1^l, y_1^l), (x_2^l, y_2^l), \cdots, (x_N^l, y_N^l)\}$. The type labels $Y = \{y_1^l, y_2^l, \cdots, y_N^l\}$ present in $D^l$ belong to $C^l$ known classes which are disjoint from the classes to discover, namely $C^l \cap C^u = \emptyset $.

\section{Method}
Our model is built on the observation that the token view and the mask view are often complementary and work well for different types. 
Thus, the core of our model is the construction of two views and how they can be utilized for co-training.

\subsection{Instance Representation}
We first describe how the relation instances are represented and then discuss the changes for event instances.
Similar to ~\cite{baldini-soares-etal-2019-matching}, in the input sentence we mark up the entity/trigger with special tokens. We use $\langle h \rangle$ and $\langle t \rangle$ for head and tail entities respectively and $\langle tgr \rangle$ for the trigger.
For each instance we have two views: the token view and the mask view. The two views share the same BERT~\cite{devlin-etal-2019-bert} encoder, but have slightly different inputs.

\paragraph{Relation Instances.}
For the token view, we embed the sentence using BERT and take the embedding for the first token in the entity (index $s$) as the entity representation.\footnote{We overload the notation a bit here and use $x$ to denote the sentence where the instance is from.} We concatenate
the representations for the head and tail entity to obtain the relation representation~\cite{baldini-soares-etal-2019-matching}.
\begin{equation}
\begin{aligned}
    \vec h &= \text{BERT}(x)[s_h] ; \vec t = \text{BERT}(x)[s_t] \\
    \vec x_1 &= [\vec h; \vec t]
\end{aligned} 
\end{equation}

For the mask view, we append a \textit{type prompt} $p_r$ to the input sentence. The type prompt is designed so that the relation type name should be fit into the \texttt{[MASK]} token position.
For relations, we use the prompt of ``{$\langle$tail$\rangle$ is the \texttt{[MASK]} of $\langle$ head $\rangle$ }'' where $\langle$tail$\rangle$ and $\langle$ head $\rangle$ are replaced by the actual head and tail entity strings for each instance.
Then we embed the sentence along with the type prompt with BERT and use the embedding for the \texttt{[MASK]} token as the relation representation. 

\begin{equation}
    \vec{x_2} = \text{BERT}(x; p_r)[s_{mask}] 
\end{equation}

\paragraph{Event Instances.}
For event instances in the token view we use the embedding for the first token in the trigger mention as the event representation.
In the mask view we use a different type prompt $p_e$: ``{$\langle$ trigger $\rangle$ is a \texttt{[MASK]} event}" where $\langle$ trigger $\rangle$ is replaced by the actual trigger.

\begin{equation}
\begin{aligned}
    \vec {x_1} & = \text{BERT}(x)[s] \\
    \vec {x_2} & = \text{BERT}(x;p_e)[s_{mask}]
\end{aligned}
\end{equation}

\subsection{Multi-view Model}
Our model consists of a shared BERT encoder, two projection networks $f$ and four classifier heads $g$ (for known types and unknown types per view, respectively). 

The projection networks map the instance representation $\vec x$ to a lower dimension space representation $\vec h$ and the classifier heads $g$ maps $\vec h$ into logits $\vec l$ corresponding to the labels.

\begin{equation}
\begin{aligned} 
    \vec h &= f(\vec x) \\
    \vec{ l^u} & = g^u(\vec h) ; \vec{ l^l} = g^l (\vec h) \\
    \vec{\hat y} &= \text{softmax}\left( [\vec{l^l}; \vec{l^u}]\right) 
\end{aligned} 
\end{equation}

For instances of known classes, we use the cross-entropy loss with label smoothing to train the network:
\begin{equation}
    \mathcal{L}^l = - \frac{1}{|D^l|}\sum_{D^l} \sum_{c=1}^C y_c \log (\hat y_c) 
\end{equation}

For instances of unknown classes, we run K-means clustering on the projection network output to assign pseudo-labels:
\begin{equation}
    \tilde{y}^u = \text{K-means}(\vec h) \in \{1,\cdots, C^u\}
\end{equation}

As the pseudo-label assignment might not align across views (cluster 1 in the token view is not the same as cluster 1 in the mask view), for each batch of instances, we further transform the cluster assignment labels into pairwise labels:
\begin{equation}
    q_{ij} = \mathbbm{1}(\tilde{y}_i = \tilde{y}_j)
\end{equation}

We compute the discrepancy between the predictions of the pair $x_i, x_j$ using the Jensen-Shannon(JS) divergence:
\begin{equation}
\small 
    d_{ij} 
    = \text{JSD} (\hat{\vec{y}}_i, \hat{\vec{y}}_j) 
    = \frac{1}{2} \left\{ \text{KL} (\hat{\vec{y}}_i || \hat{\vec{y}}_j) + \text{KL} (\hat{\vec{y}}_j || \hat{\vec{y}}_i) \right\}
\end{equation}

Then the loss function for an unlabeled pair is defined as the JS divergence if two instances are assigned to the same cluster and a hinge loss over the JS divergence if two instances are assigned to different clusters.

\begin{equation}
\small 
    \begin{aligned}
    & l(d_{ij}, q_{ij})  = q_{ij} d_{ij} + (1-q_{ij}) \max(0, \alpha - d_{ij}) \\
    & \mathcal{L}^u  = \frac{1}{\binom{|D^u|}{2}}\sum_{x_i, x_j \in D^u} \left( l(d^1_{ij}, q^2_{ij}) + l(d^2_{ij}, q^1_{ij})  \right) 
    \end{aligned} 
\end{equation}
where $d^1_{ij}$ is computed from the token view, $d^2_{ij}$ is computed from the mask view and the similarly for $q^1_{ij}$ and $q^2_{ij}$. $\alpha$ is a hyper-parameter for the hinge loss.

If a single view was used, this loss falls back to the contrastive loss term defined for unlabeled instances in \cite{Hsu2018LOC, zhao-etal-2021-relation}. 

In the training process, we observe that since the pseudo label $\tilde y$ is used as the target for the opposite view, when these two views produce very different clusters, it leads to performance oscillation over epochs.

To alleviate this issue, we add a consistency loss that encourages the predictions of the two views to be similar to each other: 
\begin{equation}
    \mathcal{L}^c = \frac{1}{|D^u|}\sum_{D^u} \text{JSD}(\hat y^1, \hat y^2) 
\end{equation}

The final loss function is a weighted sum of the aforementioned terms:
\begin{equation}
    \mathcal{L} = \mathcal{L}^l + \mathcal{L}^u + \beta \mathcal{L}^c
    \label{eq:loss}
\end{equation}
$\beta$ is a hyperparameter and empirically set to 0.2 in our experiments.

\subsection{Training Procedure}
Before we train our model with the loss function in Equation \ref{eq:loss}, we warmup our model by pre-training on the labeled data. The loss function here is simply the cross-entropy loss $\mathcal{L}_{pre} = \mathcal{L}^l$.

After pre-training, we load the weights for BERT and the projection networks $f$ to the model for further training. Note that we do not keep the weights for the known class classifier head $g^l$.

\section{Experiments}
In the following experiments, we refer to our model as \ours\ to stand for ``type abstraction''.

\begin{table}[t]
    \centering
    \small  
    \begin{tabular}{ l c c c c}
    \toprule 
    \multirow{2}{3em}{Dataset} & \multicolumn{2}{c}{Known} & \multicolumn{2}{c}{Unknown} \\
 &  \#Classes & \# Ins & \#Classes & \#Ins \\
    \midrule 
      TACRED & 31 & 23,477 & 10 & 1,996  \\
      FewRel & 64 & 44,800  & 16 & 11,200 \\
      \midrule 
      ACE-controlled & 10 & 4,089 & 23 & 1,221 \\
      ACE-end2end & 10 & 1,663 & - & 17,172 \\
      \bottomrule 
    \end{tabular}
    \caption{Statistics of the datasets used. The first two are for relation discovery and the last two datasets are used for event discovery. }
    \label{tab:dataset}
\end{table}

\subsection{Relation Discovery Setting}
\paragraph{Datasets.}
We follow RoCORE~\cite{zhao-etal-2021-relation} and evaluate our model on two relation extraction benchmark datasets: TACRED~\cite{zhang-etal-2017-position} and FewRel~\cite{han-etal-2018-fewrel}. For the TACRED dataset, 31 relation types are treated as known and 10 relation types are unknown, with the types defined in the TAC-KBP slot filling task~\cite{ji2011knowledge} \footnote{The unknown relations are \texttt{schools\_attended}, \texttt{cause\_of\_death}, \texttt{city\_of\_death}, \texttt{stateorprovince\_of\_death}, \texttt{founded}, \texttt{country\_of\_birth}, \texttt{date\_of\_birth}, \texttt{city\_of\_birth}, \texttt{charges}, \texttt{country\_of\_death}.}. Instances with the \texttt{no\_relation} label are filtered out as in \cite{zhao-etal-2021-relation}. %
For the FewRel dataset, we treat the 64 relation types in the original training set as known relations and 16 relation types in the original development set as unknown relations. 
For both datasets, we leave out 15\% of the instances of both known and unknown relation types, and we report results on the set of unknown relation instances. 
\footnote{As the data split is random, our reported numbers are not exactly the same. }

\paragraph{Baselines.}
We primarily compare with RoCORE~\cite{zhao-etal-2021-relation} and RSN~\cite{wu-etal-2019-open}. which is the state-of-the-art for relation discovery. RoCORE earns its name from their proposed ``relation-oriented clustering module'' that attempts to shape the latent space for clustering by a center loss (which pushes instances towards centroids) and a reconstruction loss.
We also compare with RSN~\cite{wu-etal-2019-open}, which learns a pairwise similarity metric between relations and transfers such a metric to unknown instances. The encoder is replaced by BERT~\cite{devlin-etal-2019-bert} for a fair comparison. RSN originally uses the Louvain algorithm~\cite{Blondel2008FastUO} for clustering, however we observe that sometimes this would lead to all instances assigned to the same cluster so we experiment with a variant using spectral clustering that takes the same graph input as Louvain. For the Louvain variant, we report the best run instead of average and deviation due to cases of clustering collapse.

\begin{table*}[t]
    \centering
    \small 
    \begin{tabular}{l | l|c c c c}
    \toprule 
   Dataset & Model      &  Acc & $B^3$ F1 & V measure & ARI  \\
   \midrule
     \multirow{3}{4em}{TACRED} & 
RSN~\cite{wu-etal-2019-open} & 
0.7645 $\pm$ 0.034& 0.7194 $\pm$ 0.036& 0.7587 $\pm$ 0.030& 0.6456 $\pm$ 0.047
\\
& RSN-spectral & 
0.7425 $\pm$ 0.041& 0.7163 $\pm$ 0.013& 0.7569 $\pm$ 0.013& 0.635 $\pm$ 0.047
\\
& ROCORE~\cite{zhao-etal-2021-relation} & 
0.8468 $\pm$ 0.059& 0.8307 $\pm$ 0.031& 0.8612 $\pm$ 0.019& 0.7867 $\pm$ 0.052
\\
& \ours & 
\textbf{0.8896} $\pm$ 0.011& \textbf{0.8535} $\pm$ 0.016& \textbf{0.8718} $\pm$ 0.017& \textbf{0.8276} $\pm$ 0.018
\\
\midrule 
\multirow{3}{4em}{FewRel} 
       & RSN~\cite{wu-etal-2019-open}
        &0.4880 & 0.4783 & 0.6718 & 0.4184  \\
& RSN-Spectral & 
0.6277 $\pm$ 0.021& 0.6306 $\pm$ 0.030& 0.7351 $\pm$ 0.020& 0.5490 $\pm$ 0.030
\\
& ROCORE~\cite{zhao-etal-2021-relation} & 
0.7801 $\pm$ 0.012& \textbf{0.7652} $\pm$ 0.025& \textbf{0.8407} $\pm$ 0.016& 0.7039 $\pm$ 0.022
\\
       & \ours  & \textbf{0.8022} $\pm$ 0.023&  0.7606 $\pm$ 0.026 & 0.8374 $\pm$ 0.018 & \textbf{0.7266} $\pm$ 0.032 \\
       \bottomrule 
    \end{tabular} 
    \caption{Relation discovery results on TACRED and FewRel. 
    Experiments are ran with 5 different seeds and we report the average score and standard deviation.  }
    \label{tab:rel_exp}
\end{table*}

\begin{table*}[t]
    \centering
    \small 
    \begin{tabular}{l l l l l }
    \toprule 
      \textbf{Model} &  \textbf{Acc} & \textbf{$B^3$ F1} & \textbf{V measure} & \textbf{ARI}   \\
      \midrule 
      \multicolumn{5}{c}{\textit{Controlled Setting}}\\
      \midrule 
    Spherical Clustering~\cite{shen-etal-2021-corpus} &  0.3830 & 0.3861 & 0.5470 & 0.2726 \\
      SS-VQ-VAE~\cite{huang-ji-2020-semi} &  0.2951 & 0.2921 & 0.4063 & 0.1242 \\
    \ours & \textbf{0.7732} $\pm$ 0.023& \textbf{0.7110} $\pm$ 0.034& \textbf{0.8028}$\pm$ 0.027& \textbf{0.6647} $\pm$ 0.038 \\
    \midrule 
    \multicolumn{5}{c}{\textit{End-to-end Setting}} \\
    \midrule 
 \ours & 0.5089 &  0.5611 & 0.7049 & 0.3629  \\
\bottomrule 
    \end{tabular}
    \caption{Event discovery results on ACE. %
    } 
    \label{tab:event_exp}
\end{table*}

\subsection{Event Discovery Setting}
\paragraph{Datasets.}
We use ACE under the processing by~\cite{lin-etal-2020-joint} for our event discovery experiments. 
We follow \cite{huang-ji-2020-semi} and set the 10 most popular event types as known types and the remaining 23 event types to be discovered.
As ACE is of relatively smaller size compared to the previous datasets used for relation discovery, we leave out 30\% of the instances for testing. Results are reported for the unknown type instances only. 

\paragraph{Controlled Setting.} 
In the controlled setting we give the models access to ground truth trigger mentions. 
We compare with the SS-VQ-VAE model from \cite{huang-ji-2020-semi} and the spherical latent clustering model from \cite{shen-etal-2021-corpus}.
As the two models originally operated on a different set of instances (sense-tagged triggers in \cite{huang-ji-2020-semi} and predicate-object pairs in \cite{shen-etal-2021-corpus}), we reimplement these methods to work with the gold-standard trigger mentions from ACE.

\paragraph{End-to-end Setting.} 
In the end-to-end setting for our system we treat all non-auxiliary verbs as candidate trigger mentions. For the 10 known types, if the annotated trigger matches with one of the candidate trigger mentions, we treat that instance as labeled. All remaining candidate triggers are treated as unknown and we set the number of unknown types $K=100$.  Under this setting, we compare with the full pipeline of ETypeClus~\cite{shen-etal-2021-corpus}.

\subsection{Metrics}
The following metrics for cluster quality evaluation are adopted: \textbf{Accuracy}, \textbf{BCubed-F1}~\cite{bagga-baldwin-1998-entity}, \textbf{V measure}~\cite{rosenberg-hirschberg-2007-v}, \textbf{Adjusted Rand Index(ARI)}~\cite{hubert1985ari}.\footnote{The implementation of BCubed is from \url{https://github.com/m-wiesner/BCUBED}, and the implementation of V measure and ARI are from the \texttt{sklearn} library.} \textbf{Accuracy} is computed by finding the maximal matching between the predicted clusters and the ground truth clusters using the Jonker-Volgenant algorithm~\cite{Crouse2016OnI2}\footnote{Implementation from \url{https://docs.scipy.org/doc/scipy/reference/generated/scipy.optimize.linear_sum_assignment.html}}. 
\subsection{Implementation Details}
We use \texttt{bert-base-uncased} as our base encoder. The projection network $f$ is implemented as a two layer MLP with dimensions 768-256-256 and ReLU activation.
The classifier heads are implemented as two layer MLPs as well,  with dimensions of 256-256-$C$, where $C$ is either the number of known types or unknown types. For additional hyperparameters, see Appendix Section \ref{sec:hyperparam}. %

\subsection{Main Results}
We present results on relation discovery in Table \ref{tab:rel_exp}. 
While all models benefit from transferring relation knowledge from known types to unknown types, RSN~\cite{wu-etal-2019-open} separates the clustering step from the representation step, so the representations are not highly optimized for clustering unlike RoCORE~\cite{zhao-etal-2021-relation} and our model.  
Compared with RoCORE, our model (1) employs a multiview representation; (2) removes the relation-oriented clustering module and 
(3) uses a simpler pretraining procedure with only known classes.
Although the training procedure is simplified, the use of both token features and mask features leads to improved effectiveness of the model. 

On the event type discovery task in Table \ref{tab:event_exp}, we show that our model has a great advantage over unsupervised methods such as spherical latent clustering model~\cite{shen-etal-2021-corpus} that does not make use of known types.
Among models that perform transfer learning, SS-VQ-VAE~\cite{huang-ji-2020-semi} does not employ a strong clustering objective over the unknown classes. %
In the end-to-end setting, our model still outperforms the previous work.
However, the gap between the end-to-end performance and the controlled performance show that extra processing on trigger might be necessary before apply this model to the wild. 
In the human evaluation Table \ref{tab:human_eval}, annotators judged 70\% of discovered clusters to be semantically coherent compared to 59\% of the clusters from the ETypeClus pipeline.

\begin{figure*}[t]
    \centering
    \includegraphics[width=\linewidth]{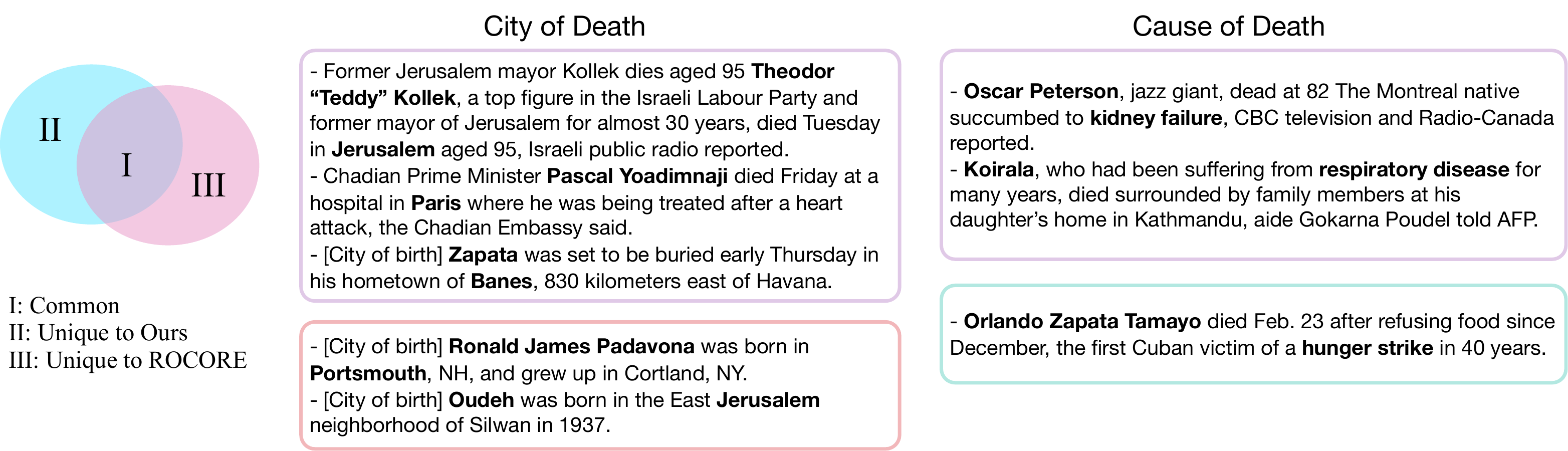}
    \caption{Comparison of predicted relation clusters on TACRED. Instances in the purple box are shared, instances in the pink box are unique to ROCORE output and instances in the blue box are unique to our model's output.}
    \label{fig:re_case_study}
\end{figure*}

\begin{table}[t]
    \centering
    \small 
    \begin{tabular}{l|l l}
    \toprule 
    \multirow{2}{3em}{Model} &  Cluster & Instance \\
    & Coherence Rate & Discernability Rate \\
     \midrule 
      ETypeClus   & 0.59 & 0.682 \\
      \ours & \textbf{0.70} & \textbf{0.725} \\
      \bottomrule 
    \end{tabular}
    \caption{Human evaluation for end-to-end event discovery on the cluster level and instance level. Reported numbers are the ratio of clusters/instances rated as ``coherent''/``discernable'' by annotators. Cohen's $\kappa=0.426$ for this binary decision process. More details about the evaluation protocol can be found in Appendix Section \ref{sec:human-eval}. }
    \label{tab:human_eval}
\end{table}

\begin{table}[ht]
    \centering
    \small 
    \begin{tabular}{l c c c }
    \toprule 
      Model   & Acc & $B^3$ & PL Acc \\
      \midrule 
    Full model    & \textbf{0.903} & \textbf{0.881} & \textbf{0.878} \\
    A:Token view only & 0.849 & 0.828 & 0.820 \\
    B:Mask view only &  0.849 & 0.832 & 0.822  \\
    C:Two branch token & 0.866 & 0.843 & 0.833 \\
    D:Two branch mask & 0.869 & 0.844 & 0.837 \\
    \bottomrule 
    \end{tabular}
    \caption{Comparison with model variations on TACRED. PL Acc is the pseudo label accuracy computed from K-means. (Results are from a single run with the same random seed.)}
    \label{tab:ablation}
\end{table}

\begin{table}[th]
    \centering
    \small 
    \begin{tabular}{l|c c}
    \toprule 
     Model   &  Acc & $B^3$  \\
     \midrule 
     Full Model & \textbf{0.903} & \textbf{0.881} \\
        w/o supervised pre-training  & 0.856 & 0.862 \\
        w/o consistency loss & 0.896 & 0.868 \\
        \bottomrule 
    \end{tabular}
    \caption{Ablation studies on the training process for TACRED.}
    \label{tab:training-ablation}
\end{table}

\begin{table*}[t]
    \centering
    \small 
    \begin{tabular}{ m{10em} m{15em} m{20em}}
    \toprule 
   Best Matched Type & Predicted Names & Instance \\
   \midrule 
   \multicolumn{3}{c }{\textit{TACRED} }\\
   \midrule 
   per:date\_of\_birth & birthday, year, years, february, month & \textbf{McNair}, born on \textbf{Dec. 14 , 1923}, in the rural Low Country of South Carolina, ...
   \\
     per:cause\_of\_death (by disease) & pmid, \textit{cord}, sign, diagnosed, cause & Palestinian leader \textbf{Abu Daoud}, who planned the daring deadly attack 
     died Saturday of \textbf{illness} ... 
     \\
     per:charges & felony, \textit{and}, \textit{anything}, cocaine, wrong  & Wen Qiang, %
     was also accused of \textbf{rape} and being unable to explain the sources of \textbf{his} assets ... \\
     \midrule 
     \multicolumn{3}{c}{\textit{ACE}} \\
     \midrule 
     Personnel:Start-Position & first, inaugural, introduced, appointed, unopposed & The ruling Millennium Democratic Party~(MDP)...
     has suffered declining popularity since President Roh Moo-Hyun \textbf{took office} in February.\\ 
     Business:Start-Org, End-Org, Merge-Org & separate, fold, new, employee, strategic & Major US insurance group AIG is 
     in a deal to \textbf{create} Japan 's sixth largest life insurer 
     \\
     Conflict:Demonstrate & street, demonstration, protest, march, picket & Chalabi staged his own \textbf{rally} yesterday to support his bid to become the next leader of Iraq. \\
      \bottomrule 
    \end{tabular}
    \caption{Predicted type names by our model. The names are sorted by frequency of appearance in top 10 predictions. We skip word pieces (starting with \#\#). Additionally, we show the top-1 instance according to prediction probability.}
    \label{tab:pred_names}
\end{table*}

\subsection{Ablation Study}
\paragraph{Different Views}
We compare our full model with several ablations of obtaining the different views as shown in Table \ref{tab:ablation}.
Variants A and B use only one view to represent the instance showing the advantage of co-training. %
We further experiment with different ways of constructing the two views.
Variant C first computes the token representation of the instance and then apply two different dropout functions to construct two views. This dropout operation can serve as task-agnostic data augmentation, which has proved to be effective for representation learning \cite{gao-etal-2021-simcse}. Variant D uses two different type abstraction prompts to construct two representations for the same instance.
Both of these variants are more effective than the single view variants but not as effective as combining the token view and the mask view. 

\paragraph{Model Design}
\label{sec:model-ablation}

In Table \ref{tab:training-ablation} we compare the performance of our full model with variants that omit the pre-training stage and the consistency loss. Pretraining the model on known types is critical to the model's final performance.
The consistency loss, while useful, does not contribute as much to the accuracy but rather alleviates the model oscillation over epochs.

\paragraph{Clustering Method}
\label{sec:clustering-ablation}
In Table \ref{tab:ablation-clustering} we experiment with different clustering methods under our framework. All implementations are from the \texttt{sklearn} library.  
For the spectral clustering variant, we use the default radial basis function (RBF) kernel to compute the affinity matrix\footnote{This is different from the spectral clustering variant of RSN, where the graph is precomputed following \cite{wu-etal-2019-open}.}, whereas for the other clustering methods we using Eulidean distance to compute the affinity matrix. This metric difference might explain why spectral clustering is underperforming. While DBSCAN and Agglomerative-Ward both achieve reasonably good performance, we observe that DBSCAN is quite sensitive to its \texttt{eps} parameter, which defines the maximum distance between two samples for one to be considered in the neighborhood of the other. In fact, this parameter needs to be set differently for different random seeds based on the distribution of the nearest neighbor distance.
In general, $k$-means clustering is both stable and efficient for our use case. 
Note that both our model and ROCORE use k-means clustering to obtain pseudo labels.

\begin{table}[t]
    \centering
    \small 
    \begin{tabular}{l|c c c }
    \toprule 
     Clustering   &  Acc & $B^3$  \\
     \midrule 
    $k$-means    & \textbf{0.9030} & \textbf{0.8806}  \\
    DBSCAN & 0.8595 & 0.8481\\ 
    Agglomerative-Ward & 0.8495 & 0.8497 \\ 
    Spectral & 0.7324 & 0.7258 \\
    \bottomrule 
    \end{tabular}
    \caption{Comparison of different clustering methods.}
    \label{tab:ablation-clustering}
\end{table}

\begin{table*}[t]
    \centering
    \small 
    \begin{tabular}{l m{38em}}
    \toprule 
   Matched Type &  Instances in Cluster \\
    \midrule 
         \multirow{4}{8em}{Personnel: Start-Position} & Condi Rice has been \textbf{chosen} by President Bush to become the new Secretary of State ...\\
        & If you were president, which national figures would you \textbf{appoint} to your cabinet and why? \\
        & Al-Douri taught international law at Baghdad University %
        before \textbf{becoming} a diplomat ... 
        \\
        & 
        Chui Sai On, who has been {\color{red} \textbf{named}~[Personnel: Nominate]} director of the SARS task force  ... \\
        \midrule 
        \multirow{3}{8em}{Conflict:Demonstrate} & Some 70 people were arrested Saturday as demonstrators clashed with police at the end of a major peace \textbf{rally} ...
        \\& Between 2,500 and 3,000 people \textbf{picketed} the CNN studios in Los Angeles ...
        \\
        & The crowd \textbf{filled} the street leading to the Kazimiya mosque in the northeast of Baghdad ... \\
        \midrule 
        \multirow{3}{8em}{Justice: Arrest-Jail} & Some 70 people were \textbf{arrested} Saturday as demonstrators clashed with police at the end of a major peace rally ...
        \\
         &Ferris disappeared from sight, and CNN has confirmed he was \textbf{taken into custody}.\\ 
        \bottomrule 
    \end{tabular}
    \caption{Predicted clusters of event instances on ACE. The triggers are marked in bold.}
    \label{tab:ace_examples}
\end{table*}

\section{Analysis}
\paragraph{Predicted Type Names.}
In Table \ref{tab:pred_names} we show the predicted type names produced by our model.
Although our model does not directly rely on such names (but rather the similarity of \texttt{[MASK]} embeddings) to perform clustering , the predictions give insights into the internal workings of the model. For example, the predicted names for the \texttt{per:cause\_of\_death} cluster are strongly related to disease.  In contrast, the following instance \textit{\textbf{Assaf Ramon}, 21, died on Sunday when the F-16 fighter \textbf{jet he was flying crashed}} was abstracted to names such as \textit{death, rotor, life, loss} and as a result, was not included as part of the cluster.

\paragraph{Relation Discovery.}
In Figure \ref{fig:re_case_study} we examine the differences in the predicted relation clusters. In the first relation \texttt{city\_of\_death}, ROCORE incorrectly merges many instances of \texttt{city\_of\_birth} into the target cluster. These two relation types not only share the same entity types of (person, city) but can also involve the exact same entities, e.g. Jerusalem. As ROCORE is primary relying on token features to make the prediction, instances with shared entities have high similarity and this propagates errors to other instances.
In the second example, we observe that both models work well in more conventional cases, but when it comes to rare values such as ``hunger strike'', only our model can correctly identify it as the \texttt{cause\_of\_death}.

\paragraph{Event Discovery.}
We show predicted clusters of event types from our algorithm under the controlled setting in Table \ref{tab:ace_examples}.
Our model is able to handle (1) \textit{diverse triggers}, e.g. ``chosen'', ``appoint'' and ``becoming'' all refer to the Start-Position event type; (2) \textit{ambiguous triggers} such as ``becoming'' and ``filled'' cannot be assigned to the event type without referring to the context; and (3) \textit{multi-word triggers}, e.g. ``take into custody'' refers to Arrest.
In the Start-Position cluster, we see a few mis-classified instances of Nominate. These two event types are similar as they both involve a person and a position/title, the difference being whether the person has already been appointed the position or not. 

\paragraph{Remaining Challenges}
\textit{Abstract types.} Relation types such as ``part\_of'',``instance\_of'' and ``same\_as'' from the FewRel dataset are highly abstract and can be associated with various types of entities.
In fact, such relations are often best dealt with separately in the context of hypernym detection~\cite{roller-etal-2018-hearst}, taxonomy construction~\cite{aly-etal-2019-every,Huang2020CoRel,chen-etal-2021-constructing} or synonym identification~\cite{Fei2019HierarchicalSynonymPrediction, shen-etal-2020-synsetexpan}.

\textit{Misaligned level of granularity.}
We observe that our automatically induced clusters are sometimes not at the same level of granularity as clusters defined by human annotation. For instance the discovered \texttt{per:cause\_of\_death} cluster is more like \texttt{per:disease\_of\_death} and the several business-related events \texttt{Start-Org}, \texttt{End-Org} and \texttt{Merge-Org} are combined into a single cluster.
This calls for models that can produce multi-level types or account for human feedback (the user can specify whether the cluster needs to be further split).

\section{Related Work}
\paragraph{Relation Type Discovery}

Early work in this direction represented relations as clusters of lexical patterns or syntactic paths ~\cite{hasegawa-etal-2004-discovering, shinyama-sekine-2006-preemptive,yao-etal-2011-structured,yao-etal-2012-unsupervised,min-etal-2012-ensemble,lopez-de-lacalle-lapata-2013-unsupervised}. %
A wave of newer methods used learned relation representations~\cite{marcheggiani-titov-2016-discrete,yu2017open,simon-etal-2019-unsupervised, wu-etal-2019-open,tran-etal-2020-revisiting, hu-etal-2020-selfore,liu-etal-2021-element,zhao-etal-2021-relation}, often defining the relation as a function of the involved entities. 
One key observation made by RSN~\cite{wu-etal-2019-open} and RoCORE~\cite{zhao-etal-2021-relation}  is the possibility of relational knowledge transfer from known relation types to new types. 
In this work, we also adopt this transfer setting and introduce a new idea of \textit{abstraction}: a relation cluster is meaningful if it aligns well with a human concept. 

\paragraph{Event Type Discovery}

Our task of event discovery is similar to the verb clustering task in SemEval 2019~\cite{qasemizadeh-etal-2019-semeval} which requires mapping verbs in context to semantic frames.\footnote{We were not able to follow this setting due to unavailable data.} 
ETypeClus~\cite{shen-etal-2021-corpus} represents events as (predicate, object) pairs and design a reconstruction-based clustering method over such P-O pairs. SS-VQ-VAE~\cite{huang-ji-2020-semi} leverages a vector quantized variational autoencoder model to utilize known types.

\section{Conclusions and Future Work}
In this paper we study the \textit{type discovery} problem:  automatically identifying and extracting new relation and event types from a given corpus. We propose to leverage \textit{type abstraction}, where the model is prompted to name the type, as an alternative view of the data.
We design a co-training framework, 
and demonstrate that our framework works favorably in both relation and event type discovery settings.
Currently we have assumed that the new types are disjoint to the old types and the model operates similarly to a transfer learning setting. While the model can be easily extended to handle both new types and old types, more analysis might be needed in this direction. One potential direction would be to explore a continual learning setting, where new types could emerge periodically. %

\section{Limitations}
In this paper we studied datasets that are English and mostly in the newswire genre. Although our method is not strictly restricted to English, the design of the type-inducing prompt will require some prior knowledge about the target language. 

For both relation and event type discovery, the model requires the input of candidate entities pairs or triggers. As shown in Table \ref{tab:event_exp}, there is a large gap in model performance between the controlled setting and the end-to-end setting (although this could be partially attributed to incomplete annotation and our simple candidate extraction process). This would limit the model's application in the real world and we believe this should be the focus of future research.

\section{Ethical Considerations}
\indent \textit{Intended use.}
The model introduced in this paper is intended to be used for exploratory analysis of datasets.
For instance, when presented with a new corpus, the model can be used to extract clusters of new event types that can then be judged by human annotators and used as a basis for developing an event ontology and event extraction system.

\indent \textit{Biases.} The model does not perform any filtering of its input. 
If the input corpus contains mentions of discriminatory or offensive language, the model will be unaware and will likely surface such issues in its output.

\section*{Acknowledgements}
We would like to thank Yiqing Xie and Jiaming Shen for early discussions about the project idea. We also appreciate the help of Jinfeng Xiao, Yizhu Jiao and Yunyi Zhang on the evaluation.  
This research was supported by US DARPA KAIROS Program No. FA8750-19-2-1004.
Any opinions, findings, and conclusions or recommendations expressed herein are those of the authors and do not necessarily represent the views, either expressed or implied, of DARPA or the U.S. Government.

\bibliography{anthology,custom}
\bibliographystyle{acl_natbib}

\appendix
\label{sec:appendix}
\section{Experiment Details}
\label{sec:hyperparam}
We use an effective batch size of 32 (among $\{8, 16, 32, 64\}$) and train with an initial learning rate of $5e-5$ (among $\{1e-5, 3e-5, 5e-5, 1e-5\}$) using the AdamW optimizer and a linear schedule.
The model is pretrained for 3 epochs for initialization and then further trained for 30 epochs on TACRED/ACE and 20 epochs on FewRel.  
For the hyperparameters in our model, we set the margin for the hinge loss $\alpha=2$ following ~\cite{Hsu2018LOC}. We show some additional tuning results in Table \ref{tab:hyperparameter_alpha}. The weight for the consistency loss $\beta=0.2$ was tuned from $\{0.1, 0.2, 0.5\}$. 
 We tuned our hyperparameters on TACRED based on accuracy and applied them to FewRel and ACE.

Our models are trained on a single Nvidia RTX A6000 GPU. A single run on TACRED takes 2 hours, a run on FewRel takes 2.5 hours and a run on ACE takes 40 minutes. Our model has 111M parameters (110M are from bert-base).

\begin{table}[t]
    \centering
    \small 
    \begin{tabular}{l|c c}
    \toprule 
    $\alpha$     &   Acc & $B^3$ \\
    \midrule 
      0.5   & 0.9063 & 0.8841 \\
      1 & 0.8696 & 0.8493 \\
      2 & 0.9030 &  0.8806 \\
      5 & 0.9063 & 0.8842 \\
      \bottomrule 
    \end{tabular}
    \caption{Tuning the hinge loss margin $\alpha$ on TACRED.}
    \label{tab:hyperparameter_alpha}
\end{table}

\section{Varying Cluster Number $K$}
\label{sec:vary-cluster}

\begin{figure}
    \centering
    \includegraphics[width=\linewidth]{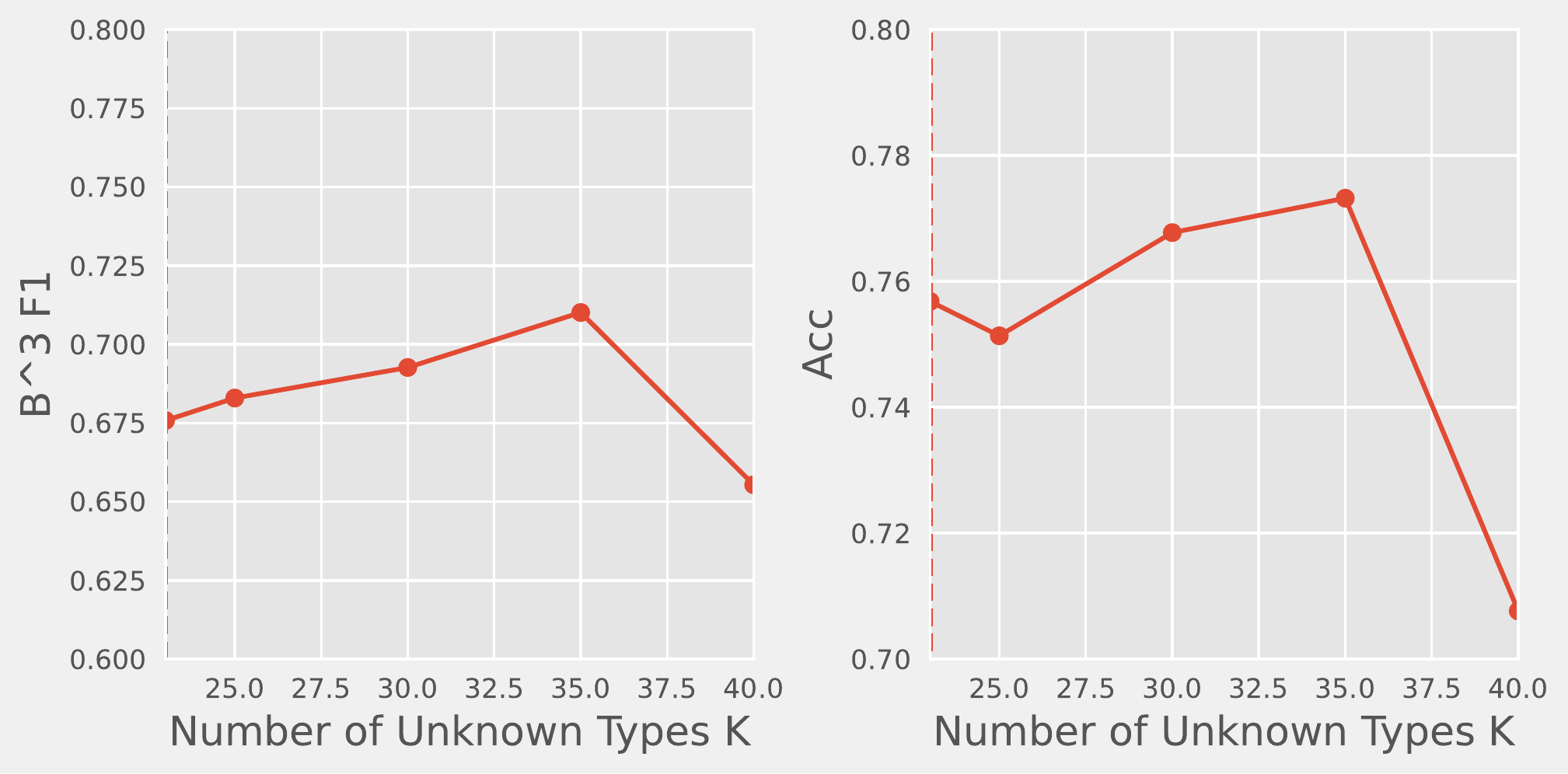}
    \caption{Performance of type discovery on ACE with varying cluster number $K$. The ground truth number of clusters $K=23$.}
    \label{fig:ace_var}
\end{figure}

\begin{figure}
    \centering
    \includegraphics[width=\linewidth]{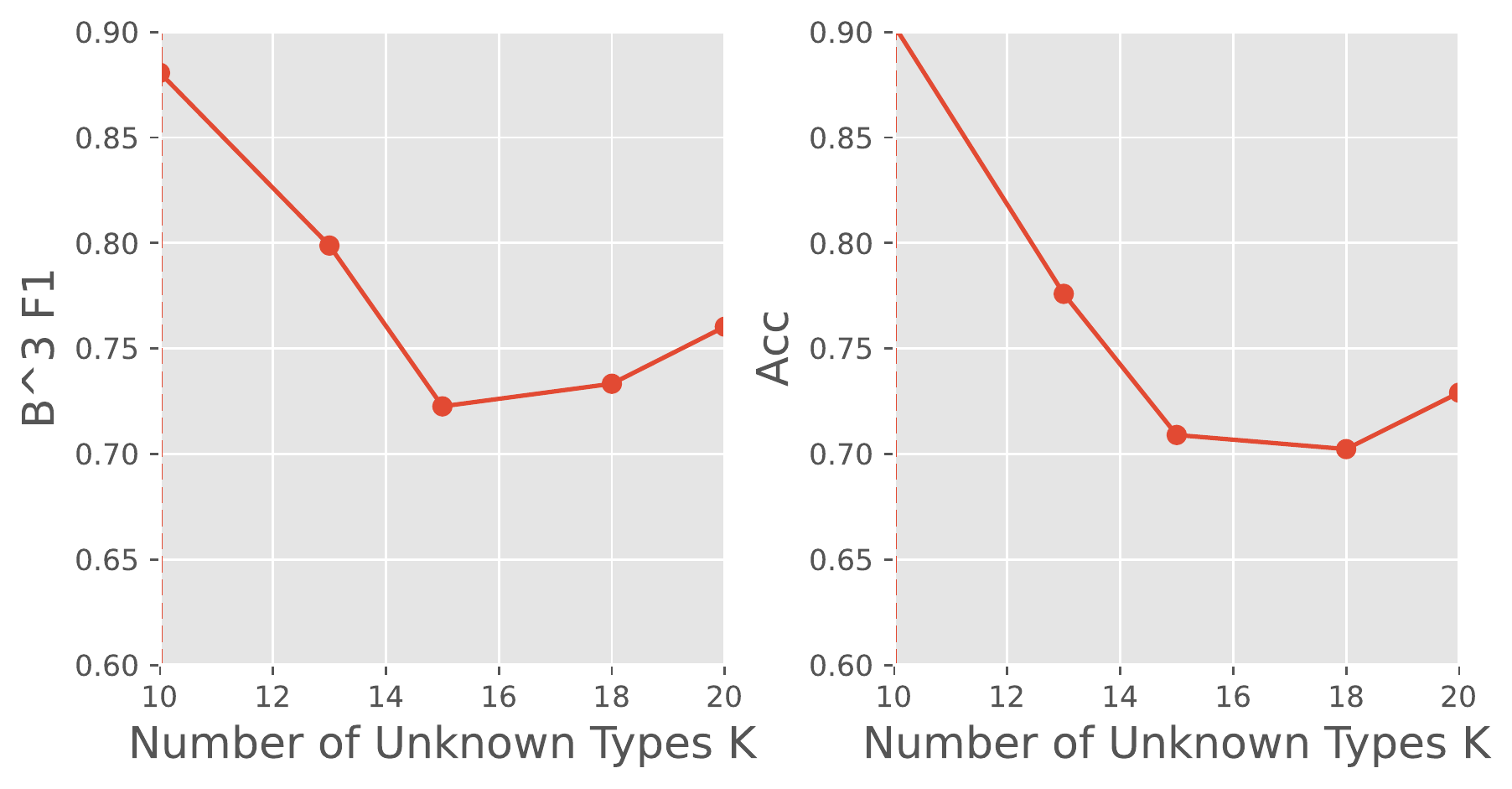}
    \caption{Performance of type discovery on TACRED with varying cluster number $K$. The ground truth number of clusters $K=10$.}
    \label{fig:tacred_var}
\end{figure}

In Figure \ref{fig:ace_var} and \ref{fig:tacred_var} we show how the model's performance changes with different specified number of unknown types $K$. Generally speaking, $K$ will impact the granularity of the discovered types. 
On the ACE dataset, a slightly larger number of $K$ will lead to improved performance. At $K=35$, the model is able to separate \texttt{Business:End-Org} from \texttt{Business:Merge-Org} which were originally mixed at $K*=23$.
On TACRED, though, $K*=10$ seems to be the optimal value, and a larger $K=20$ would result in \texttt{per:cause\_of\_death} being split into subcategories of disease, homicide, accident and \texttt{per: charges} being split into subcategories of violent (e.g. murder) and non-violent (e.g. espionage).

\section{Human Evaluation Protocol}
\label{sec:human-eval}
We evaluate the end-to-end results for event discovery from both the cluster level and instance level.
For each cluster, we present the top 10 and bottom 10 instances and ask annotators if this cluster is meaningful and relevant to the corpus. 
For instance-level evaluation, we ask the annotator whether an instance belongs to a set of candidate instances or not. This set of candidate instances is either sampled from the same predicted cluster or randomly selected from other clusters with 50\% probability. 

\begin{figure*}
    \centering
    \includegraphics[width=\linewidth]{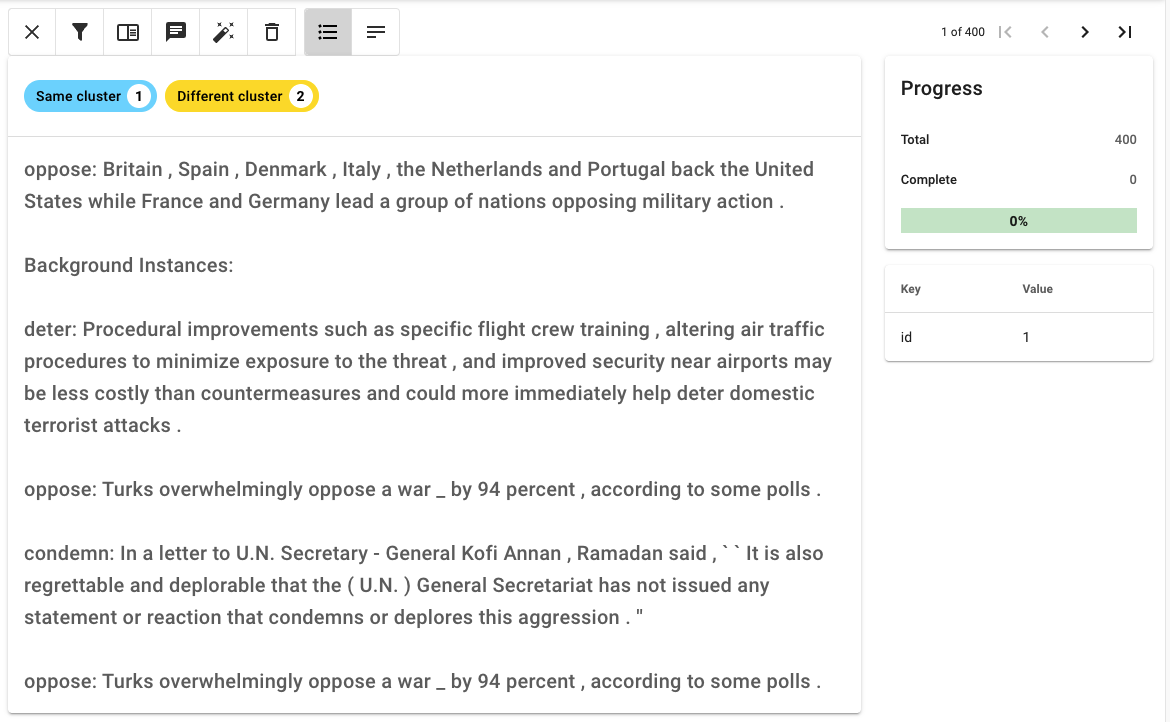}
    \caption{An example of the evaluation interface presented to annotators.}
    \label{fig:interface}
\end{figure*}

\section{End-to-end Event Discovery Case Study}
In Table \ref{tab:e2e_ace} we show the results of our model along with ETypeClus under the end-to-end setting. 
The pipeline of ETypeClus converts predicate mentions into predicate-object~(P-O) pairs, selects salient P-O pair then clusters such salient pairs. 
As a result, the output clusters do not cover infrequent triggers such as ``swinging'' and ``siphoning'' and the clusters themselves are often tied together by shared predicates or shared objects (\texttt{establish state}, \texttt{establish administration} and \texttt{endorse administration}).
Our model, on the other hand, operates directly on predicate mentions, allowing us to identify events with infrequent triggers and events with named entity or pronoun objects as in ``set up EasyJet'' and ``blame each other''.

\begin{table*}[ht]
    \centering
    \small 
    \begin{tabular}{c|m{10em} | m{4em} m{28em} }
    \toprule 
        Event Type  & ETypeClus & \multicolumn{2}{c}{Ours}  \\
         & \textbf{Predicate-Obj} & \textbf{Predicate} & \textbf{Mentions} \\ 
        \midrule
        \multirow{4}{4em}{Transaction} & \multirow{4}{10em}{\underline{return-1 piece}, sell-3 cookie, sell-3 commercial, buy-0 pudding, sell-0 park, \underline{build-0 housing}, sell-5 share} & 
        sell & 
        The program allows Iraq to \textbf{sell} unlimited quantities of oil to buy food \\
        & & buying & They're basically \textbf{buying} future medical care throughout their lives \\
        & & swinging & Motorola and Texas Instruments both in the chips base \textbf{swinging} to profits \\
        & & siphoning & He had also been accused of \textbf{siphoning} millions of dollars from Project Coast to finance a lavish, globe-trotting lifestyle \\
        \midrule 
        \multirow{4}{4em}{Create} & \multirow{4}{10em}{build-2 blog, establish-0 country, form-2 group, \underline{endorse-1 administration},
        \underline{incorporate-0 blog}, establish-0 state,
        establish-0 administration} 
        & create & Major US insurance group AIG is in the final stage of talks ... in a deal to \textbf{create} Japan 's sixth largest life insurer  \\
        & & produce & The electricity that Enron \textbf{produced} was so exorbitant that the government decided it was cheaper not to buy electricity \\
        & & set (up) & EasyCinema founder Stelios Haji - Ioannou , who \textbf{set up} easyJet in 1995 \\
        & & \underline{pass} & U.S. Ambassador John Negroponte was asked whether the United States would withdraw the resolution if it didn't have the votes to \textbf{pass} it \\
        \midrule 
        \multirow{4}{4em}{Oppose} & \multirow{4}{10em}{\underline{maintain-1 innocence}, \underline{plead-2 conspiracy}, denounce-0 move, reject-0 change, rid-0 move, oppose-0 move, announce-2 creation, denounce-1 presence} &  rejecting & the flight attendants came in with a close vote \textbf{rejecting} these concessions  \\
        & & opposed & 78 of 100 people surveyed \textbf{opposed} the military action in Iraq  \\
        & & blamed & A summit ... had been planned for Wednesday but was postponed, according to Israeli and Palestinian officials , who \textbf{blamed} each other for the delay. \\
        & & objected & Russia \textbf{objected} to World Bank rules that required monitoring of patients receiving medication \\
        \bottomrule 
    \end{tabular}
    \caption{Discovered type clusters in the end-to-end setting on ACE. The event type names were manually assigned based on the cluster content. The predicate mentions are in bold. The questionable assignments are underlined.}
    \label{tab:e2e_ace}
\end{table*}

\end{document}